\documentclass[11pt,a4paper]{article}
\pdfoutput=1
\usepackage[hyperref]{acl2021}
\usepackage{times}
\usepackage{latexsym}

\usepackage{microtype}

\aclfinalcopy 
\def\aclpaperid{*} 

\usepackage{graphicx}
\usepackage{svg}
\usepackage{amsmath,amssymb}
\usepackage{here}
\usepackage{url}
\usepackage{comment}
\usepackage{booktabs}
\usepackage{multirow}
\usepackage{tabularx}
\usepackage{color}
\usepackage{tcolorbox}
\usepackage{CJK}
\usepackage{fancybox}
\usepackage{colortbl}
\usepackage{soul}
\tcbset{width=0.9\textwidth,boxrule=0pt,colback=red,arc=0pt,auto outer arc,left=0pt,right=0pt,boxsep=5pt}

\newcommand{\squaddu}[1]{SQuAD$^{\rm Du}_{\rm #1}$}

\newcommand{\Parskip}{-3pt}
\newcommand{\Itemsep}{3pt}
\newcommand{\Itemindent}{0pt}
\newcommand{\Leftmargin}{-11pt}

\title{Improving the Robustness of QA Models to Challenge Sets\\
with Variational Question-Answer Pair Generation}

\author{Kazutoshi Shinoda$^{\dagger \ddagger}$ ~~ Saku Sugawara$^\ddagger$ ~~ Akiko Aizawa$^{\dagger \ddagger}$\\
    $^\dagger$The University of Tokyo\\
    $^\ddagger$National Institute of Informatics\\
    {\tt shinoda@is.s.u-tokyo.ac.jp} \\
    {\tt \{saku,aizawa\}@nii.ac.jp} \\
}

\date{}

\begin{document}
\maketitle
\begin{abstract}
Question answering (QA) models for reading comprehension have achieved human-level accuracy on in-distribution test sets.
However, they have been demonstrated to lack robustness to challenge sets, whose distribution is different from that of training sets.
Existing data augmentation methods mitigate this problem by simply augmenting training sets with synthetic examples sampled from the same distribution as the challenge sets.
However, these methods assume that the distribution of a challenge set is known a priori, making them less applicable to unseen challenge sets.
In this study, we focus on question-answer pair generation (QAG) to mitigate this problem.
While most existing QAG methods aim to improve the quality of synthetic examples,
we conjecture that diversity-promoting QAG can mitigate the sparsity of training sets and lead to better robustness.
We present a variational QAG model that generates multiple diverse QA pairs from a paragraph.
Our experiments show that our method can improve the accuracy of 12 challenge sets, as well as the in-distribution accuracy.\footnote{Our code and data are available at \url{https://github.com/KazutoshiShinoda/VQAG}.}
\end{abstract}

\section{Introduction}

Machine reading comprehension has gained significant attention in the NLP community, whose goal is to devise systems that can answer questions about given documents \cite{Rajpurkar16,Trischler17,Joshi17}.
Such systems usually use neural models, which require a substantial number of question-answer (QA) pairs for training.
To reduce the considerable manual cost of dataset creation, there has been a resurgence of studies on automatic QA pair generation (QAG), consisting of a pipeline of answer extraction (AE) and question generation (QG), to augment question answering (QA) datasets \cite{Yang17,Du18,Subramanian18,Alberti19}.

For the downstream QA task, most existing studies have evaluated QAG methods using a test set from the same distribution as a training set \cite{Yang17,Zhang19,Liu-et-al-2020-Asking-Questions}.
However, when a QA model is evaluated only on an in-distribution test set, it is difficult to verify that the model is not exploiting unintended biases in a dataset \cite{Geirhos2020}.
Exploiting an unintended bias can degrade the robustness of a QA model, which is problematic in real-world applications.
For example, recent studies have observed that a QA model does not generalize to other QA datasets \cite{yogatama2019learning,talmor-berant-2019-multiqa,sen-saffari-2020-models}.
Other studies have found a lack of robustness to challenge sets, such as paraphrased questions \cite{gan-ng-2019-improving}, questions with low lexical overlap \cite{sugawara-etal-2018-makes}, and questions that include noise \cite{ravichander-etal-2021-noiseqa}.

While existing studies have proposed data augmentation methods targeting a particular challenge set, they are only effective at the expense of the in-distribution accuracy \cite{gan-ng-2019-improving,ribeiro-etal-2019-red,ravichander-etal-2021-noiseqa}.
These methods assume that the target distribution is given a priori.
However, identifying the type of samples that a QA model cannot handle in advance is difficult in real-world applications.

We conjecture that increasing the diversity of a training set with data augmentation, rather than augmenting QA pairs similar to the original training set, can improve the robustness of QA models.
Poor diversity in QA datasets has been shown to result in the poor robustness of QA models \cite{Lewis19,geva-etal-2019-modeling,ko-etal-2020-look}, supporting our hypothesis.
To this end, we propose a variational QAG model (VQAG).
We introduce two independent latent random variables into our model to learn the two one-to-many relationships in AE and QG by utilizing neural variational inference \cite{vae}.
Incorporating the randomness of these two latent variables enables our model to generate diverse answers and questions separately.
We also study the effect of controlling the Kullback--Leibler (KL) term in the variational lower bound for mitigating the posterior collapse issue \cite{Bowman16}, where the model ignores latent variables and generates outputs that are almost the same.
We evaluate our approach on 12 challenge sets that are unseen during training to assess the improved robustness of the QA model.

In summary, our contributions are three-fold:
\begin{itemize}
\setlength{\parskip}{\Parskip}
\setlength{\itemsep}{\Itemsep}
\setlength{\itemindent}{\Itemindent}
\setlength{\leftmargin}{\Leftmargin}
\item We propose a variational question-answer pair generation model with explicit KL control to generate significantly diverse answers and questions.
\item We construct synthetic QA datasets using our model to boost the QA performance in an in-distribution test set, achieving comparable scores with existing QAG methods.
\item We discover that our method achieves meaningful improvements in unseen challenge sets, which are further boosted using a simple ensemble method.
\end{itemize}


\section{Related Work}
\subsection{Answer Extraction}
AE aims to extract question-worthy phrases, which are worth being asked about, from each textual context without looking at the questions.
AE has been performed mainly in two ways: rule-based and neural methods.
\citet{Yang17} extracted candidate phrases using rule-based methods such as named entity recognition (NER).
However, not all the named entities, noun phrases, verb phrases, adjectives, or clauses in the given documents are used as gold answer spans.
As such, these rule-based methods are likely to extract many trivial phrases.

Therefore, there have been studies on training neural models to identify question-worthy phrases.
\citet{Du18} framed AE as a sequence labeling task and used BiLSTM-CRF \cite{Huang15}.
\citet{Subramanian18} treated the positions of answers as a sequence and used a pointer network \cite{ptrnet}.
\citet{Wang19} used a pointer network and Match-LSTM \cite{WangJiang16,WangJiang17}.
\citet{Alberti19} made use of pretrained BERT \cite{bert}.

However, these neural AE models are trained with maximum likelihood estimation; that is, each model is optimized to produce an answer set closest to the gold answers.
In contrast, our model incorporates a latent random variable and is trained by maximizing the lower bound of the likelihood to extract diverse answers.
In this study, we assume that there should be question-worthy phrases that are not used as the gold answers in a manually created dataset.
We aim to extract such phrases.

\subsection{Question Generation}
Traditionally, QG was studied using rule-based methods \cite{Mostow09,Heilman10,Lindberg13,Labutov15}.
After \citet{Du17} proposed a neural sequence-to-sequence model \cite{Sutskever14} for QG, neural models that take context and answer as inputs have started to be used to improve question quality with attention \cite{Bahdanau14} and copying \cite{Gulcehre16,Gu16} mechanisms.
Most works focused on generating relevant questions from context-answer pairs \cite{Zhou18,Song18,Zhao18,Xingwu18,Kim18,Liu19,Qiu19}.
These works showed the importance of answers as input features for QG.
Other works studied predicting question types \cite{Zhou19,Kang19}, modeling a structured answer-relevant relation \cite{Li19}, and refining generated questions \cite{Nema19}.
To further improve question quality, policy gradient techniques have been used \cite{Yuan17,Yang17,Yao18,Kumar18}.
\citet{Dong19} used a pretrained language model.

The diversity of questions has been tackled using variational attention \citep{Bahuleyan18}, a conditional variational autoencoder (CVAE) \citep{Yao18}, and top p nucleus sampling \citep{sultan-etal-2020-importance}.
Our study is different from these studies wherein we study QAG by introducing variational methods into both AE and QG.
\citet{lee-etal-2020-generating} is the closest to our study in terms of the modeling choice.
While \citet{lee-etal-2020-generating} introduced an information-maximizing term to improve the consistency of QA pairs, our study uniquely controls the diversity by explicitly controlling KL values.

Despite the potential of data augmentation with QAG to mitigate the sparsity of QA datasets and avoid overfitting, not much is known about the robustness of QA models reinforced with QAG to more challenging test sets.
We comprehensively evaluate QAG methods on challenging QA test sets, such as hard questions \cite{sugawara-etal-2018-makes}, implications \cite{ribeiro-etal-2019-red}, and paraphrased questions \cite{gan-ng-2019-improving}.

\subsection{Variational Autoencoder}
The variational autoencoder (VAE) \cite{vae} is a deep generative model consisting of a neural encoder (inference model) and decoder (generative model).
The encoder learns to map from an observed variable to a latent random variable and the decoder works vice versa.
The techniques of VAE have been widely applied to NLP tasks such as text generation \cite{Bowman16}, machine translation \cite{Zhang16}, and sequence labeling \cite{Chen18}.

The CVAE is an extension of the VAE, in which the distribution of a latent variable is explicitly conditioned on certain variables and enables generation processes to be more diverse than a VAE \cite{cvae-poem,cvae-dialog,Shen17}.
The CVAE is trained by maximizing the variational lower bound of the log likelihood.


\section{VQAG: Variational Question-Answer Pair Generation Model}

\subsection{Problem Definition}

Our problem is to generate QA pairs from textual contexts.
We focus on extractive QA in which an answer is a text span in context.
We use $c$, $q$, and $a$ to represent the context, question, and answer, respectively.
We assume that every QA pair is sampled independently given a context. Thus, the problem is defined as maximizing the conditional log likelihood ${\log p(q, a|c)}$ averaged over all samples in a dataset.

\subsection{Variational Lower Bound with Explicit KL Control}
\label{sec:objective}

Generating questions and answers from different latent spaces makes sense because multiple questions can be created from a context-answer pair and multiple answer spans can be extracted from a context.
Thus, we introduce two independent latent random variables to assign the roles of diversifying AE and QG to $z$ and $y$, respectively.

VAEs often suffer from \textit{posterior collapse}, where the model learns to ignore latent variables and generates outputs that are almost the same \cite{Bowman16}.
Many approaches have been proposed to mitigate this issue, such as weakening the generators \cite{Bowman16,Yang-vae17,Semeniuta17}, or modifying the objective functions \cite{wae,infovae,beta-vae}.

To mitigate this problem, we use a variant of the modified $\beta$-VAE \cite{beta-vae} proposed by \citet{Burgess18}, which uses two hyperparameters to control the KL terms.
Our modified objective function is:
\begin{align}
\label{eq:modified-elbo}
\mathcal{L} & = \mathbb{E}_{q_{\phi}(z, y|q, a, c)}[\log p_{\theta} (q|y, a, c) \notag\\
& + \log p_{\theta} (a|z, c)]\notag\\
& - |D_{\rm KL}(q_{\phi}(z|a, c)||p_{\theta}(z|c)) - {\rm C_a}| \notag\\
& - |D_{\rm KL}(q_{\phi}(y|q, c)||p_{\theta}(y|c)) - {\rm C_q}|,
\end{align}
\noindent where $D_{\rm KL}$ is the KL divergence, $\theta$ ($\phi$) is the parameters of the generative (inference) model, and ${\rm C_a}, {\rm C_q} \geq 0$.
See Appendix \ref{app:derivation} for the derivation of the objective.
Tuning ${\rm C_a}$ and ${\rm C_q}$ was enough to regularize the KL terms in our case (see Appendix \ref{app:modeling-capacity}).
${\rm C_a}$ and ${\rm C_q}$ can explicitly control the KL values because the KL terms are forced to get closer to these values during training.
We mathematically show that the KL control can be interpreted as controlling the conditional mutual information $I(Z; A|C)$ and $I(Y; Q|C)$.
This is the major difference between our model and \citet{lee-etal-2020-generating}, where $I(Q; A)$ is maximized to improve consistency of QA pairs.\footnote{The upper cases represent the random variables.}
See Appendix \ref{app:interpretation} for the mathematical interpretation.

\begin{figure}[t]
 \begin{center}
    \includegraphics[clip,width=\linewidth]{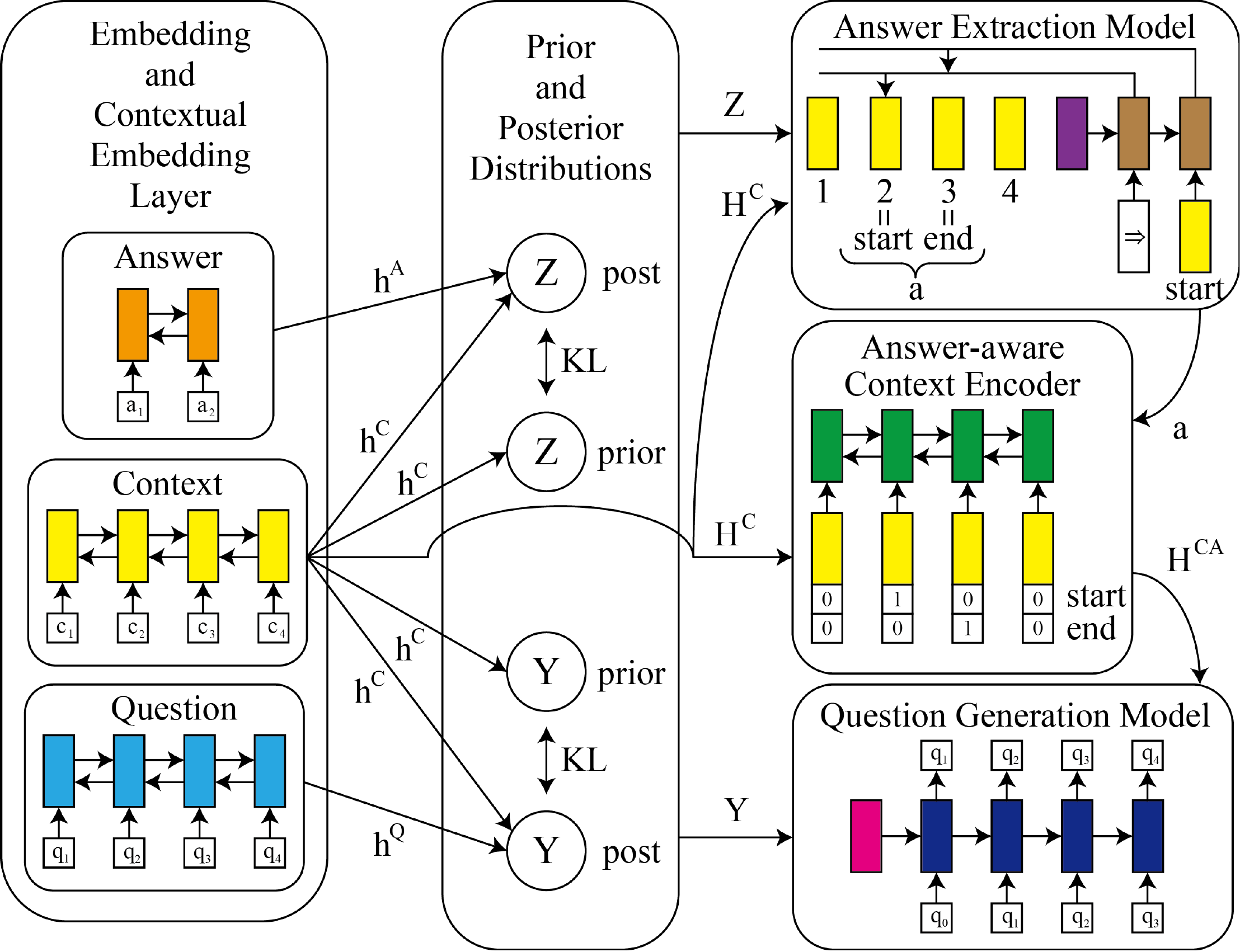}
    \caption{Overview of the model architecture. The latent variables $z$ and $y$ are sampled from the posteriors when computing the variational lower bound and from the priors during generation. See {\S}\ref{sec:model} for the details.}
    \label{fig:overview}
  \end{center}
\end{figure}

\subsection{Model Architecture}
\label{sec:model}

An overview of VQAG is given in Figure \ref{fig:overview}.
We denote $c_i$, $q_i$, and $a_i$ as the $i$-th word in context, question, and answer, respectively. See Appendix \ref{app:architechture} for the details of the implementation.

\paragraph{Embedding and Contextual Embedding Layer}
First, in the embedding layer, the $i$-th word, $w_i$, of a sequence of length $L$ is simultaneously converted into word- and character-level embedding vectors
by using a CNN based on \citet{Kim14}.
Then, we concatenate the embedding vectors.
After that, we pass the embedding vectors to the contextual embedding layer consisting of bidirectional LSTMs (BiLSTM).
We obtain $H \in\mathbb{R}^{L \times 2d}$, which is the concatenated outputs from the LSTMs in each direction at each time step, and $h \in\mathbb{R}^{2d}$, which is the concatenated last hidden state vectors in each direction.
The superscripts of the outputs $H$ and $h$ shown in Figure \ref{fig:overview}
indicate where they come from. $C$, $Q$, and $A$ denote the context, question, and answer, respectively.

\paragraph{Prior and Posterior Distributions}
Following \citet{cvae-dialog}, we hypothesize that the prior and and posterior distributions of the latent variables follow multivariate Gaussian distributions with diagonal covariance.
The mean $\mu$ and log variance $\log\sigma^2$ of these prior and posterior distributions of $z$ and $y$ are computed with linear transformation from $h^C$, $h^A$, and $h^Q$.
Next, latent variable $z$ and $y$ are obtained using the reparameterization trick \cite{vae}.
Then, $z$ and $y$ are passed to the AE and QG models, respectively.
$z$ and $y$ are sampled from the posteriors during training and the priors during testing.

\paragraph{Answer Extraction Model}
We regard AE as two-step autoregressive decoding, i.e.,
$p(a|c)=p(c_{start}|c)p(c_{end}|c_{start}, c)$,
that predicts the start and end positions of an answer span in this order.
For AE, we modify a pointer network \cite{ptrnet} to take as input the initial hidden state computed from linear transformation from $z$, which in the end diversifies AE by learning the mappings from $z$ to $a$.
We use an LSTM as a decoder and compute attention scores over $H^C$.

\paragraph{Answer-aware Context Encoder}
To compute answer-aware context information for QG, we use another BiLSTM.
We concatenate $H^C$ and one hot vectors of start and end positions of answer as in Figure \ref{fig:overview}, which are fed to the BiLSTM.
We obtain $H^{CA} \in \mathbb{R}^{L \times 2d}$, which is the concatenated outputs from the LSTMs in each direction.
$H^{CA}$ is used as the source for attention and copying in QG.

\paragraph{Question Generation Model}
For QG, we modify an LSTM decoder with attention and copying mechanisms to take as input the initial hidden state computed from linear transformation from $y$, which in the end diversifies QG.
At each time step, the probability distribution of generating words from vocabulary $P_{v}(q_i)$ is computed using attention \cite{Bahdanau14}, and the probability distributions of copying words \cite{Gulcehre16,Gu16} from context $P_{c}(c_j)$ are computed using attention.
In parallel, the switching probability $p_s$ is linearly estimated from the hidden state vector.
Lastly, we compute the probability of $q_i$ as:
\begin{align}
p(q_i) = p_{s} P_{v} (q_i)  + (1 - p_{s})
\sum_{j:c_j={q_i}} P_{c}(c_j).
\end{align}

%
%

\section{Experiments and Results}
\subsection{Dataset}
We used SQuAD v1.1 \cite{Rajpurkar16}, a large scale QA dataset consisting of documents collected from Wikipedia and 100k QA pairs created by crowdworkers, as a source dataset for QAG.
Answers to questions in SQuAD can be extracted from textual contexts.
Since the SQuAD test set has not been released, we use the split of the dataset, SQuAD-Du \cite{Du17}, where the original training set is split into the training set (\squaddu{train}) and the test set (\squaddu{test}), and the original development set is used as the dev set (\squaddu{dev}).
The sizes of \squaddu{train}, \squaddu{dev}, and \squaddu{test} are 75,722, 10,570, and 11,877, respectively. See Appendix \ref{app:training-details} for the training details of VQAG.


\begin{table}[t]
\setlength{\tabcolsep}{2pt}
\small
\centering
\begin{tabular}{lrrlrrlr}
\toprule
\multicolumn{1}{c}{} & \multicolumn{5}{c}{Relevance} & & \multicolumn{1}{l}{Diversity} \\
\cmidrule{2-6}\cmidrule{8-8}
\multicolumn{1}{l}{} & \multicolumn{2}{c}{Precision} &  & \multicolumn{2}{c}{Recall} & & \multicolumn{1}{c}{\multirow{2}{*}{Dist}} \\
\cmidrule{2-3}\cmidrule{5-6}
\multicolumn{1}{c}{} & \multicolumn{1}{c}{Prop.}  & \multicolumn{1}{c}{Exact} &  & \multicolumn{1}{c}{Prop.} & \multicolumn{1}{c}{Exact} &  &  \\
\midrule
NER & 34.44 & 19.61 &  & 64.60 & 45.39 &  & 30.0k \\
HarQG & 45.96 &  33.90 &  & 41.05 &  28.37 &  & - \\
InfoHCVAE & 31.59 & 16.18 &  & 78.75 & 59.32 &  & 70.1k\\
\midrule
VQAG & \multicolumn{1}{l}{} & \multicolumn{1}{l}{} &  & \multicolumn{1}{l}{} & \multicolumn{1}{l}{} &  & \multicolumn{1}{l}{} \\
~~${\rm C_a=0}$ & \textbf{58.39} &  \textbf{47.15} &  & 21.82 & 16.38 &  & 3.1k \\
~~${\rm C_a=5}$ & 30.16 & 13.41 &  & \textbf{83.13} & \textbf{60.88} &  & 71.2k \\
~~${\rm C_a=20}$ & 21.95 & 5.75 &  & 72.26 & 42.15 &  & \textbf{103.3k} \\
\bottomrule
\end{tabular}
\caption{Results of answer extraction on \squaddu{test}. Prop.: Proportional Overlap, Exact: Exact Match, Dist: the number of distinct context-answer pairs. $\rm C_a$ is the hyperparameter in Eq. \ref{eq:modified-elbo}.}
\label{tb:ae}
\end{table}


\begin{table}[tbp]
\setlength{\tabcolsep}{2pt}
\small
\centering
\begin{tabular}{lrrrrrrr}
\toprule
\multicolumn{1}{l}{} & \multicolumn{3}{c}{Relevance}                                                  & \multicolumn{1}{l}{}      & \multicolumn{3}{c}{Diversity}                                             \\
\cmidrule{2-4}\cmidrule{6-8}
\multicolumn{1}{l}{} & \multicolumn{1}{c}{B1-R} & \multicolumn{1}{c}{ME-R} & \multicolumn{1}{c}{RL-R} & \multicolumn{1}{c}{Token} & \multicolumn{1}{c}{D1} & \multicolumn{1}{c}{E4} & \multicolumn{1}{c}{SB4} \\\midrule
SemQG          & \textbf{62.32}           & \textbf{36.77}           & \textbf{62.87}           & 7.0M                      & 15.8k                  & 18.28                  & 91.44                   \\
VQAG &&&&&&&\\
~~${\rm C_q=0}$                    & 35.57                     & 18.31                    & 33.92                    & 7.6M                      & 14.4k                  & 17.33                  & 97.61                   \\
~~${\rm C_q=5}$                    & 44.19                     & 25.84                    & 45.18                    & 11.5M                     & 19.0k                  & 19.71                  & 82.59                   \\
~~${\rm C_q=20}$                   & 48.19                    & 25.29                    & 48.26                    & 4.9M                      & \textbf{22.4k}         & \textbf{19.72}         & \textbf{44.41}      \\\bottomrule
\end{tabular}
\caption{Results of answer-aware question generation on \squaddu{test}.
50 questions for each context-answer pair are generated and evaluated to assess their diversity. B1-R, ME-R, RL-R is the recall of BLEU-1, METEOR, and Rouge-L. D1: Dist-1, E4: Ent-4, SB4: Self-BLEU-4. $\rm C_q$ is the hyperparameter in Eq. \ref{eq:modified-elbo}.
}
\label{tb:qg}
\end{table}

\subsection{Answer Extraction}
\label{sec:exp-ae}
First, we conducted the AE experiment, where inputs were contexts and outputs were a set of multiple answer spans.
The objective of this experiment is to measure the diversity and the extent to which our extracted answers cover the ground truths.
We also study the effect of ${\rm C_a}$ in Eq. \ref{eq:modified-elbo}.

\paragraph{Metrics} To measure the accuracy of multi-span extraction, we computed Proportional Overlap (Prop.) and Exact Match (Exact) metrics \citep{Breck07,Johansson10,Du18} for each pair of extracted and ground truth answer spans, and then we report their precision and recall.\footnote{We exclude Binary Overlap, which assigns higher scores to systems that extract the entire input context and is not a reliable metric as \citet{Breck07} discussed.}
Prop. is proportional to the amount of overlap between two phrases.
Our models extracted 50 answers from each context.
To measure the diversity, we defined a Dist score, which is the the total number of distinct context-answer pairs.

\paragraph{Baselines} We used three baselines: named entity recognition (NER), Harvesting QG (HarQG) \cite{Du18}, and InfoHCVAE \cite{lee-etal-2020-generating}.
For NER, we used spaCy \cite{spacy2}.
For HarQG, we directly copied the scores from \citet{Du18}.
For InfoHCVAE, we trained the model on the training set, and extracted 50 answers randomly from each context for a fair comparison.

\paragraph{Result} Table \ref{tb:ae} shows the result.
While we tested various values of ${\rm C_a}$ ranging from 0 to 100, we only report the selected values here for brevity.
When using ${\rm C_a}$ larger than 20, the scores did not get improved.
Our model with ${\rm C_a=5}$ performed the best in terms of the recall scores, while surpassing the diversity of NER.
The highest Dist scores did not occur together with the highest recall scores.
When ${\rm C_a}$ is 0, the Dist score is fairly low.
This implies the posterior collapse issue, though the precision scores are the best.
We assert that low precision scores do not necessarily mean poor performance in our experiment because even the original test set does not cover all the valid answer spans.

\subsection{Answer-aware Question Generation}
\label{sec:exp-qg}

We also conducted answer-aware QG experiments where the contexts and ground truth answer spans were the inputs to assess diversity and relevance to the gold questions.

\paragraph{Metrics}~ To evaluate the diversity of the generated questions, our models generated 50 questions from each context-answer pair.
We reported the recall scores (denoted as ``-R'') of BLEU-1 (B1), METEOR (ME), and ROUGE-L (RL) per reference question.
We do not report precision scores here because our motivation is to improve diversity.
To measure diversity, we reported ${\rm Dist}$-1 (D1), Entropy-4 (E4) \cite{Serban17,Zhang18}, and Self-BLEU-4 (SB4) \cite{texygen}.\footnote{We computed ${\rm Dist}$-1 following the definition of \citet{Jingjing18}, wherein ${\rm Dist}$-1 is the number of distinct unigrams. ${\rm Dist}$-1 is often defined as the ratio of distinct unigrams \cite{Li16} but this is not fair when the number of generated sentences differs among models, so we did not use this. SB4 was calculated per 50 questions generated from each input.}

\paragraph{Baselines} We compared our models with SemQG \cite{Zhang19}.\footnote{We reran the ELMo+QPP\&QAP model, which is available at \url{https://github.com/ZhangShiyue/QGforQA}.}
We used diverse beam search \cite{divbeam}, sampled the top 50 questions per answer from SemQG, and used them to calculate the metrics as the baseline for a fair comparison

\paragraph{Result} The results in Table \ref{tb:qg} show that our model can improve diversity while degrading the recall scores compared to SemQG.
Using ${\rm C_q}$ larger than 20 did not lead to improved diversity.
More detailed analysis of ${\rm C_a}$ and ${\rm C_q}$ is provided in Appendix \ref{app:ae-qg}.

\begin{table*}[t]
\small
\centering
\scalebox{0.8}{
\begin{tabular}{ll}
\toprule
\multicolumn{2}{c}{
{\setlength{\fboxsep}{0pt}\colorbox{white!0}{\parbox{
19cm
}{
\colorbox{red!44.11764705882353}{\strut beyoncé} \colorbox{red!26.47058823529412}{\strut 's} \colorbox{red!20.588235294117645}{\strut vocal} \colorbox{red!20.588235294117645}{\strut range} \colorbox{red!38.23529411764706}{\strut spans} \Ovalbox{ 
\Ovalbox{ 
\colorbox{red!100.0}{\strut four} 
}\colorbox{red!70.58823529411765}{\strut octaves} 
}\colorbox{red!17.647058823529413}{\strut .} \Ovalbox{ 
\colorbox{red!47.05882352941176}{\strut jody} \colorbox{red!44.11764705882353}{\strut rosen} 
}\colorbox{red!8.823529411764707}{\strut highlights} \colorbox{red!2.941176470588235}{\strut her} \colorbox{red!2.941176470588235}{\strut tone} \colorbox{red!2.941176470588235}{\strut and} \colorbox{red!2.941176470588235}{\strut timbre} \colorbox{red!2.941176470588235}{\strut as} \colorbox{red!2.941176470588235}{\strut particularly} \colorbox{red!11.76470588235294}{\strut distinctive} \colorbox{red!5.88235294117647}{\strut ,} \colorbox{red!2.941176470588235}{\strut describing} \colorbox{red!2.941176470588235}{\strut her} \colorbox{red!2.941176470588235}{\strut voice} \colorbox{red!2.941176470588235}{\strut as} \colorbox{red!2.941176470588235}{\strut "} \colorbox{red!35.294117647058826}{\strut one} \colorbox{red!32.35294117647059}{\strut of} \colorbox{red!32.35294117647059}{\strut the} \colorbox{red!32.35294117647059}{\strut most} \colorbox{red!32.35294117647059}{\strut compelling} \colorbox{red!32.35294117647059}{\strut instruments} \colorbox{red!35.294117647058826}{\strut in} \colorbox{red!52.94117647058824}{\strut popular} \colorbox{red!52.94117647058824}{\strut music} \colorbox{red!8.823529411764707}{\strut "} \colorbox{red!0.0}{\strut .} \colorbox{red!0.0}{\strut while} \colorbox{red!0.0}{\strut another} \colorbox{red!0.0}{\strut critic} \colorbox{red!0.0}{\strut says} \colorbox{red!0.0}{\strut she} \colorbox{red!0.0}{\strut is} \colorbox{red!0.0}{\strut a} \colorbox{red!5.88235294117647}{\strut "} \colorbox{red!32.35294117647059}{\strut vocal} \colorbox{red!17.647058823529413}{\strut acrobat} \colorbox{red!5.88235294117647}{\strut ,} \colorbox{red!0.0}{\strut being} \colorbox{red!0.0}{\strut able} \colorbox{red!0.0}{\strut to} \colorbox{red!5.88235294117647}{\strut sing} \colorbox{red!5.88235294117647}{\strut long} \colorbox{red!2.941176470588235}{\strut and} \colorbox{red!8.823529411764707}{\strut complex} \colorbox{red!5.88235294117647}{\strut melismas} \colorbox{red!5.88235294117647}{\strut and} \colorbox{red!5.88235294117647}{\strut vocal} \colorbox{red!5.88235294117647}{\strut runs} \colorbox{red!5.88235294117647}{\strut effortlessly} \colorbox{red!0.0}{\strut ,} \colorbox{red!0.0}{\strut and} \colorbox{red!0.0}{\strut in} \colorbox{red!0.0}{\strut key} \colorbox{red!0.0}{\strut .} \Ovalbox{ 
\colorbox{red!5.88235294117647}{\strut her} \colorbox{red!14.705882352941178}{\strut vocal} \colorbox{red!14.705882352941178}{\strut abilities} 
}\colorbox{red!0.0}{\strut mean} \colorbox{red!26.47058823529412}{\strut she} \colorbox{red!26.47058823529412}{\strut is} \colorbox{red!26.47058823529412}{\strut identified} \colorbox{red!23.52941176470588}{\strut as} \colorbox{red!32.35294117647059}{\strut the} \colorbox{red!47.05882352941176}{\strut centerpiece} \colorbox{red!47.05882352941176}{\strut of} \colorbox{red!73.52941176470588}{\strut destiny} \colorbox{red!64.70588235294117}{\strut 's} \colorbox{red!64.70588235294117}{\strut child} \colorbox{red!17.647058823529413}{\strut .} \Ovalbox{ 
\colorbox{red!29.411764705882355}{\strut the} \colorbox{red!44.11764705882353}{\strut daily} \colorbox{red!61.76470588235294}{\strut mail} 
}\colorbox{red!32.35294117647059}{\strut calls} \colorbox{red!32.35294117647059}{\strut beyoncé} \colorbox{red!32.35294117647059}{\strut 's} \colorbox{red!29.411764705882355}{\strut voice} \colorbox{red!29.411764705882355}{\strut "} \Ovalbox{ 
\colorbox{red!32.35294117647059}{\strut versatile} 
}\colorbox{red!29.411764705882355}{\strut "} \colorbox{red!29.411764705882355}{\strut ,} \colorbox{red!29.411764705882355}{\strut capable} \colorbox{red!29.411764705882355}{\strut of} \colorbox{red!29.411764705882355}{\strut exploring} \colorbox{red!29.411764705882355}{\strut power} \colorbox{red!29.411764705882355}{\strut ballads} \colorbox{red!29.411764705882355}{\strut ,} \colorbox{red!32.35294117647059}{\strut soul} \colorbox{red!32.35294117647059}{\strut ,} \colorbox{red!32.35294117647059}{\strut rock} \colorbox{red!32.35294117647059}{\strut belting} \colorbox{red!32.35294117647059}{\strut ,} \colorbox{red!32.35294117647059}{\strut operatic} \colorbox{red!32.35294117647059}{\strut flourishes} \colorbox{red!32.35294117647059}{\strut ,} \colorbox{red!32.35294117647059}{\strut and} \Ovalbox{ 
\colorbox{red!32.35294117647059}{\strut hip} \colorbox{red!32.35294117647059}{\strut hop} 
}\colorbox{red!20.588235294117645}{\strut .} \colorbox{red!82.35294117647058}{\strut jon} \colorbox{red!82.35294117647058}{\strut pareles} \colorbox{red!23.52941176470588}{\strut of} \colorbox{red!23.52941176470588}{\strut the} \colorbox{red!23.52941176470588}{\strut new} \colorbox{red!23.52941176470588}{\strut york} \colorbox{red!23.52941176470588}{\strut times} \colorbox{red!17.647058823529413}{\strut commented} \colorbox{red!11.76470588235294}{\strut that} \colorbox{red!11.76470588235294}{\strut her} \colorbox{red!11.76470588235294}{\strut voice} \colorbox{red!8.823529411764707}{\strut is} \colorbox{red!8.823529411764707}{\strut "} \colorbox{red!14.705882352941178}{\strut velvety} \colorbox{red!14.705882352941178}{\strut yet} \Ovalbox{ 
\colorbox{red!14.705882352941178}{\strut tart} 
}\colorbox{red!8.823529411764707}{\strut ,} \colorbox{red!8.823529411764707}{\strut with} \colorbox{red!8.823529411764707}{\strut an} \colorbox{red!14.705882352941178}{\strut insistent} \colorbox{red!14.705882352941178}{\strut flutter} \colorbox{red!14.705882352941178}{\strut and} \colorbox{red!23.52941176470588}{\strut reserves} \colorbox{red!17.647058823529413}{\strut of} \colorbox{red!26.47058823529412}{\strut soul} \colorbox{red!26.47058823529412}{\strut belting} \colorbox{red!5.88235294117647}{\strut "} \colorbox{red!0.0}{\strut .}
}}}
}
\\
\midrule
Q: how can one find her vocal abilities in key music ? & A: she is identified as the centerpiece of destiny 's child\\
Q: how many octaves spans beyoncé 's vocal range ? & A: spans four\\
Q: how many octaves 's vocal range spans the beyoncé hop vocal range ? & A: four\\
Q: who commented that her voice is tart yet tart ? & A: jon pareles\\
\bottomrule
\end{tabular}
}
\caption{Heatmap of extracted answer spans and generated samples using our model.
The darker the color is, the more often the word is extracted.
The phrases surrounded by black boxes are the ground truth answers in SQuAD.}
\label{tb:heatmap}
\end{table*}

\subsection{Synthetic Dataset Construction}
We created three synthetic QA datasets, denoted as ${ \mathcal{D}_{5,5}}$, ${\mathcal{D}_{20,20}}$, and ${\mathcal{D}_{5,20}}$, using VQAG with the different configurations, $({\rm C_a}, {\rm C_q}) = (5, 5), (20, 20), (5, 20)$ respectively.
These configurations are chosen based on the recall-based metrics and diversity scores in the AE and QG results.

VQAG generated 50 QA pairs from each paragraph in \squaddu{train} to construct each ${\mathcal{D}}$.
It is generally known that VAEs generate diverse but low-quality data unlike GANs.
We used heuristics to filter out low-quality generated QA pairs, dropping questions that are longer than 20 words or shorter than 5 words and answers that are longer than 10 words, keeping questions that have at least one interrogative word, and removing n-gram repetition in questions.
While some existing works used the BERT QA model or an entailment model as a data filter \cite{Alberti19,Zhang19,Liu-et-al-2020-Asking-Questions},
our heuristics are enough to obtain improvement in the downstream QA task as shown in \S\ref{sec:qa}.
Some samples in our datasets are given in Table \ref{tb:heatmap}, showing that the diverse QA pairs are generated.
See Appendix \ref{app:latent-interpolation} to see how VQAG maps the latent variables to the QA pairs.

\subsection{Human Evaluation}
\label{sec:human}
We assess the quality of the synthetic QA pairs by conducting human evaluation on Amazon Mechanical Turk.
For human evaluation, we randomly chose 200 samples from synthetic QA pairs generated by \citet{Zhang19} and our model with $({\rm C_a}, {\rm C_q}) = (5, 5), (20, 20)$ from the paragraphs in \squaddu{test}.
We also chose 100 samples from \squaddu{test}.
In addition to the three items proposed by \citet{Liu-et-al-2020-Asking-Questions}, we asked annotators if an given answer is important, i.e., it is worth being asked about.
We showed the workers a triple (passage, question, answer) and asked them to answer the four questions shown in Table \ref{tb:human}.
See Appendix \ref{app:human} for the details.
We report the responses obtained using the majority vote.

\begin{table}[]
\setlength{\tabcolsep}{2pt}
\small
\scalebox{0.9}{
\begin{tabular}{c|c|rrrr}
\toprule
\multicolumn{2}{c|}{Experiments}                                                                     & \multicolumn{1}{c}{SemQG} &
\multicolumn{2}{c}{
\begin{tabular}{cc}
     \multicolumn{2}{c}{$({\rm C_a},{\rm C_q})=$}\\
      $(5,5)$ & $(20, 20)$
\end{tabular}
}
& \multicolumn{1}{c}{SQuAD} \\
\midrule
\multirow{3}{*}{\begin{tabular}[c]{@{}c@{}}Question is\\ well-formed\end{tabular}} & No             & 2.9\%                        & 23.1\%                 & 27.8\%                  & 2.3\%                     \\
                                                                                   & Understandable & 34.5\%                       & 16.0\%                 & 17.0\%                  & 10.5\%                    \\
                                                                                   & Yes            & 62.6\%                       & 60.9\%                 & 55.1\%                  & 87.2\%                    \\\midrule
\multirow{2}{*}{\begin{tabular}[c]{@{}c@{}}Question is\\ relevant\end{tabular}}    & No             & 2.5\%                        & 9.5\%                  & 11.5\%                  & 4.0\%                     \\
                                                                                   & Yes            & 97.5\%                       & 90.5\%                 & 88.5\%                  & 96.0\%                    \\\midrule
\multirow{3}{*}{\begin{tabular}[c]{@{}c@{}}Answer is\\ correct\end{tabular}}       & No             & 2.8\%                        & 28.8\%                 & 30.5\%                  & 7.5\%                     \\
                                                                                   & Partially      & 21.8\%                       & 28.1\%                 & 26.6\%                  & 11.8\%                    \\
                                                                                   & Yes            & 75.4\%                       & 43.2\%                 & 42.9\%                  & 80.6\%                    \\\midrule
\multirow{2}{*}{\begin{tabular}[c]{@{}c@{}}Answer is\\ important\end{tabular}}     & No             & 1.5\%                        & 10.0\%                 & 5.0\%                   & 6.0\%                     \\
                                                                                   & Yes            & 98.5\%                       & 90.0\%                 & 95.0\%                  & 94.0\%\\\bottomrule
\end{tabular}
}
\caption{Human evaluation of the quality of QA pairs. $\rm C_a$ and $\rm C_q$ are the hyperparameters in Eq. \ref{eq:modified-elbo}.}
\label{tb:human}
\end{table}

According to the results in Table \ref{tb:human}, nearly 25\% of our questions are not understandable or meaningful, and 30\% of our answers are incorrect for the generated questions.
This result indicates that our synthetic datasets contain a considerable number of noisy QA pairs in these two aspects.
However, 90\% of the generated questions are relevant to the passages, and 90\% of the answers extracted by our models are question-worthy.
As we will verify in \S\ref{sec:qa}, our noisy but diverse synthetic datasets are effective in enhancing the QA performance in the in- and out-of-distribution test sets.

\subsection{Question Answering}
\label{sec:qa}
We evaluated QAG methods on the downstream QA task.
We evaluated our method on 12 challenge sets in addition to the in-distribution test set.

\subsubsection{Baselines}
We compared our method with the following baselines.
\begin{itemize}
\setlength{\parskip}{\Parskip}
\setlength{\itemsep}{\Itemsep}
\setlength{\itemindent}{\Itemindent}
\setlength{\leftmargin}{\Leftmargin}
    \item \textbf{\squaddu{train}} BERT-base model trained on \squaddu{train} without data augmentation.
    \item \textbf{HarQG} \cite{Du18} uses neural AE and QG models and generates over one million QA pairs from top ranking Wikipedia articles not included in SQuAD. We used the publicly available dataset.\footnote{\url{https://github.com/xinyadu/harvestingQA}}
    \item \textbf{SemQG} \cite{Zhang19} uses reinforcement learning to generate more SQuAD-like questions. We reran the trained model, and generated questions from the same context-answer pairs as HarQG.
    \item \textbf{InfoHCVAE} \cite{lee-etal-2020-generating} uses a variational QAG model with an information-maximizing term. We trained this model\footnote{\url{https://github.com/seanie12/Info-HCVAE}} on \squaddu{train}, and then generated 50 QA pairs from each context in \squaddu{train} for a fair comparison with VQAG.
\end{itemize}

\subsubsection{Training Details}
We trained pretrained BERT-base models \cite{bert} on each synthetic dataset, and then fine-tuned it on \squaddu{train}.
We adopted this procedure following existing data augmentation approach for QA \cite{dhingra-etal-2018-simple,Zhang19}.
In our study, the order in which our synthetic datasets $\mathcal{D}$ were given to a QA model was tuned on the dev set.

We used the Hugging Face's implementation of BERT \cite{wolf-etal-2020-transformers}.
We used Adam \cite{kingma2014adam} with epsilon as 1e-8 for the optimizer.
The batch size was 32.
In both the pretraining and fine-tuning procedure, the learning rate decreased linearly from 3e-5 to zero.
We conducted the training for one epoch using a synthetic dataset and two epochs using the original training set.

In addition to the performance of \textit{Single} models, we reported the performance of \textit{Ensemble} models, where the output probabilities of three different QA models are simply averaged.
In practice, the top 20 candidate answer spans predicted by each QA model were used for the final prediction.

\begin{table*}[htbp]
\setlength{\tabcolsep}{4pt}
\renewcommand{\arraystretch}{1.2}
\small
\centering
\begin{tabular}{@{}cc|
>{\columncolor[HTML]{EFEFEF}}c c
>{\columncolor[HTML]{EFEFEF}}c c
>{\columncolor[HTML]{EFEFEF}}c c
>{\columncolor[HTML]{EFEFEF}}c c
>{\columncolor[HTML]{EFEFEF}}c c
>{\columncolor[HTML]{EFEFEF}}c c
>{\columncolor[HTML]{EFEFEF}}c }
\toprule
& & \multicolumn{13}{c}{Challenge Sets} \\
         & Training Data (Size) & \squaddu{test} & News & NQ   & Quo & Para & APara & Hard & Imp  & Add  & AddO & MT   & ASR  & KB   \\
\hline
\parbox[t]{2mm}{\multirow{5}{*}{\rotatebox[origin=c]{90}{\textit{Single}}}}   & \squaddu{train} (76k) & 83.5 & 49.2 & 67.7 & 30.1 & 85.7 & 50.2 & 75.6 & 64.7 & 62.9 & 71.8 & 79.7 & 67.5 & 80.1 \\
 & +HarQG (1,205k) & \textcolor{red}{\textit{83.3}} & \textcolor{red}{\textit{48.5}} & \textcolor{red}{\textit{66.2}} & 31.3 & \textcolor{red}{\textit{85.2}} & 56.5 & \textcolor{red}{\textit{73.0}} & \textcolor{red}{\textit{63.5}} & 65.1 & 73.1 & \textcolor{red}{\textit{78.6}} & 70.0 & 80.3 \\
 & +SemQG (1,204k) & 84.7 & 50.5 & 69.8 & \textbf{\textit{34.5}} & 86.2 & 51.8 & \textcolor{red}{\textit{75.0}} & 65.1 & \textbf{\textit{66.5}} & 74.3 & \textcolor{red}{\textit{79.5}} & 71.0 & 80.7 \\
 & +InfoHCVAE (824k) & \textbf{84.8} & \textbf{51.3} & \textbf{71.2} & 33.8 & \textcolor{red}{85.6} & 53.3 & \textbf{77.7} & 64.8 & 66.1 & \textbf{74.5} & \textbf{81.3} & \textbf{71.6} & \textbf{82.8} \\
 & +VQAG (432k) & 84.5 & 49.2 & 70.1 & 32.0 & \textbf{86.7} & \textbf{59.0} & 76.1 & \textbf{66.3} & 64.8 & 73.9 & 79.9 & 70.5 & 81.1 \\
\hline
\parbox[t]{2mm}{\multirow{5}{*}{\rotatebox[origin=c]{90}{\textit{Ensemble}}}}
& \{\squaddu{train}\}*3 & 84.2 & 50.4 & 69.4 & 31.3 & 86.4 & 53.2 & 76.6 & 65.7 & 63.6 & 72.6 & 80.3 & 68.7 & 81.2 \\
 & \{+SemQG\}*3 & 85.5 & 51.8 & 71.3 & \textbf{35.1} & 87.5 & 57.8 & 78.2 & 66.5 & 67.0 & 75.1 & 80.8 & 72.9 & 82.5 \\
 & \{+InfoHCVAE\}*3 & 85.3 & 52.0 & \textbf{72.2} & 34.0 & 88.0 & 56.9 & \textbf{79.0} & 65.7 & \textbf{67.7} & \textbf{75.9} & 81.4 & 73.1 & \textbf{83.2} \\
 & \{+VQAG\}*3 & 84.9 & 50.9 & 70.1 & 32.3 & \textbf{88.1} & \textbf{58.6} & 77.3 & \textbf{67.5} & 64.9 & 73.9 & 80.8 & 71.2 & 81.6 \\
 & \{+Sem,+Info,+V\} & \textbf{85.8} & \textbf{52.1} & 72.0 & 34.2 & 88.0 & 55.1 & 78.8 & 67.0 & 66.3 & 74.7 & \textbf{82.2} & \textbf{73.5} & 83.0 \\\hline \hline
 \multicolumn{2}{c|}{\textit{If challenge set is known}} & - & 62.9 & 83.0 & 66.9 & 88.6 & 73.9 & - & - & - & - & 80.8 & 75.9 & 82.6 \\
\bottomrule
\end{tabular}
\caption{QA performance (F1 score) on \squaddu{test} and the 12 challenge sets. The abbreviations of the challenge sets are explained in \S\ref{sec:qa}.
Curly brackets denote an ensemble of different models (e.g., \{+VQAG\}*3 denotes the ensemble of three QA models, trained with different random seeds after data augmentation with VQAG). The best scores for each of the \textit{Single} and \textit{Ensemble} models are \textbf{boldfaced}. The degraded scores compared to the no data augmentation baseline (the 1st line) are in \textcolor{red}{\textit{red}}. Sem: SemQG, Info: InfoHCVAE, V: VQAG.}
\label{tb:robustness}
\end{table*}

\subsubsection{Challenge Sets}
We assessed the robustness of the QA models to the following 12 challenge sets, as well as \squaddu{test}.
\begin{itemize}
\setlength{\parskip}{\Parskip}
\setlength{\itemsep}{\Itemsep}
\setlength{\itemindent}{\Itemindent}
\setlength{\leftmargin}{\Leftmargin}
    \item \textbf{NewsQA (News)} \cite{Trischler17}: 5,166 QA pairs created from CNN articles by crowdworkers, transformed into the SQuAD format following \citet{sen-saffari-2020-models}.
    \item \textbf{Natural Questions (NQ)} \cite{kwiatkowski-etal-2019-natural}: 2,356 questions from real users for Wikipedia articles. We reframed NQ as extractive QA by using long answers in NQ as contexts following \citet{sen-saffari-2020-models}.\footnote{We used answerable questions for NewsQA and NQ provided by \citet{sen-saffari-2020-models}. We did not use the MRQA shared task version as \citet{lee-etal-2020-generating} did.}
    \item \textbf{Non-Adversarial Paraphrased Test Set (Para)} \cite{gan-ng-2019-improving}: 1,062 questions paraphrased with slight perturbations from SQuAD using a trained paraphrased model.
    \item \textbf{Adversarial Paraphrased Test Set (APara)} \cite{gan-ng-2019-improving}: 56 questions manually paraphrased using context words near a confusing answer from SQuAD.
    \item \textbf{Hard Subset (Hard)} \cite{sugawara-etal-2018-makes}: A subset of the SQuAD dev set, which consists of 1,661 questions that require less word matching and more knowledge inference and multiple sentence reasoning.
    \item \textbf{Implications (Imp)} \cite{ribeiro-etal-2019-red}: 13,371 QA pairs automatically derived from the SQuAD dev set with a linguistic rule-based method.\footnote{For example, ``Q: \textit{Who died in 1285?} A: \textit{Zhenjin}'' is derived from ``Q: \textit{When did Zhenjin die?} A: \textit{1285}''}
    \item \textbf{AddSent (Add) \& AddOneSent (AddO)} \cite{jia-liang-2017-adversarial}: Adversarial SQuAD dataset created using handcrafted rules designed for fooling a QA model. The sizes of Add and AddO are 3,560 and 1,787, respectively.
    \item \textbf{Quoref (Quo)} \cite{dasigi-etal-2019-quoref}: 2,418 questions requiring coreference resolution created by humans. We used the dev set.
    \item \textbf{Natural Machine Translation Noise (MT)} \cite{ravichander-etal-2021-noiseqa}: A subset of NoiseQA, consisting of 1,190 English translated questions produced by Google's commercial translation system from the XQuAD dataset \cite{artetxe-etal-2020-cross}. This creation introduces naturally occurring noise caused by machine translation.
    \item \textbf{Natural Automatic Speech Recognition Noise (ASR)} \cite{ravichander-etal-2021-noiseqa}: Another subset of NoiseQA, consisting of 1,190 questions that include automatic speech recognition error.
    \item \textbf{Natural Keyboard Noise (KB)} \cite{ravichander-etal-2021-noiseqa}: Another subset of NoiseQA, consisting of 1,190 questions that include natural character-level typos introduced by typing questions on a keyboard.
\end{itemize}
\noindent These challenge sets enable us to evaluate the QA models' robustness to other domain corpora, variations in questions, adversarial examples, and noise that may occur in real-world applications.

\subsubsection{Results}
The overall results are given in Table \ref{tb:robustness}.
First, we discovered that the QA model without data augmentation degraded the performance on the 12 challenge sets, showing a lack of the robustness to the natural and adversarial distribution shifts in contexts, questions, and answers.\footnote{The score on Para—85.7 F1 is degraded when compared to the score on the SQuAD dev set—87.9 F1, which is the source for creating Para. This means the lack of robustness to paraphrased questions as shown in \citet{gan-ng-2019-improving}.}

With data augmentation using QAG, the in-distribution scores were generally improved, except for HarQG.
In the \textit{Single} model setting on the challenge sets, SemQG achieved the best performance on Quo and Add.
InfoHCVAE achieved the best performance on News, NQ, Hard, AddO, MT, ASR, and KB.
VQAG achieved the best performance on Para, APara, and Imp.
These results imply that different QAG methods have different benefits.
In the \textit{Ensemble} setting, taking the best of the three, the scores on \squaddu{test}, News, MT, and ASR were further improved with \{+Sem,+Info,+V\}.

We also attached scores that are obtained \textit{if challenge set is known} in Table \ref{tb:robustness}; that is, natural or synthetic samples from the same distributions as the challenge sets are available during training.
For News, NQ, and Quo, we trained the BERT-base model on the corresponding training sets, which are annotated by humans.
For paraphrased questions (Para, APara) and NoiseQA (MT, ASR, and KB), the scores were taken from \citet{gan-ng-2019-improving} and \citet{ravichander-etal-2021-noiseqa}, respectively.
These scores can be considered as the upper bounds.
In NoiseQA, the QAG methods consistently improved the scores, even though they were not designed for the noise.
This may be because the lack of quality in synthetic datasets, as shown in Table \ref{tb:human}, unintentionally improved the robustness to the noise.
However, the most significant performance gap ($>$ 30 F1) between the upper bound and the no data augmentation baseline was observed in Quo.
This result indicates that a QA model does not acquire coreference resolution from SQuAD, even though approximately 18\% of SQuAD questions require coreference resolution \cite{sugawara-etal-2018-makes}.
The QAG methods mitigated this gap to some extent, but there is a significant room for improvement.

The improvement in NQ is generally more prominent than that in News.
This may be because both SQuAD and NQ contain paragraphs in Wikipedia.
Utilizing unlabeled documents in domains such as news articles may improve the generalization to other domains such as News.


In paraphrased questions \cite{gan-ng-2019-improving}, implications \cite{ribeiro-etal-2019-red}, and NoiseQA \cite{ravichander-etal-2021-noiseqa}, augmenting questions that are similar to the corresponding challenge sets, that is, generating paraphrases, implications, and questions including the noise, successfully improved the robustness to these perturbations.
While these methods slightly degraded or maintained the in-distribution score, we showed that QAG methods are less likely to exhibit a trade-off between the in- and out-of-distribution accuracies.
Notably, VQAG did not degrade the scores on all the 12 challenge sets while improving the in-distribution score.
In contrast, SemQG degraded the scores on Hard and MT, and InfoHCVAE degraded the score on Para.
This property of VQAG may be because it can significantly improve the diversity by combining the different configurations.

Moreover, the size of the synthetic dataset created by VQAG was the smallest among the QAG methods as shown in Table \ref{tb:robustness}.
If the diversity is assured sufficiently, significantly increasing the quantity may not be necessary.
In Add and AddO, we showed that the QAG methods consistently improved adversarial robustness, which has not been studied in the QAG literature.

\begin{table}[t]
\centering
\small
\begin{tabular}{c|rr}
\toprule
Training Data (Size) & \multicolumn{1}{c}{EM}  & \multicolumn{1}{c}{F1} \\
\midrule
VQAG (432k)        & 81.49                   & 88.61                  \\
$-{\mathcal{D}_{5,5}}$ (251k)        & 81.04                   & 88.39                  \\
$-{\mathcal{D}_{5,20}}$ (113k)   & 81.00                   & 88.48                 \\
$-{\mathcal{D}_{20,20}}$ (68k)        & 81.14                   & 88.52                  \\
\bottomrule
\end{tabular}
\caption{Ablation study on \squaddu{dev}. Each synthetic dataset is shown to be useful to improve the scores.}
\label{tb:ablation}
\end{table}

\subsubsection{Analysis}

To assess the usefulness of each dataset $\mathcal{D}$ in VQAG, we conducted an ablation study.
As shown in Table \ref{tb:ablation}, each dataset $\mathcal{D}$ has meaningful effect on the performance.
This result implies that creating more synthetic datasets using different configurations may further improve the performance.

To understand the differences in each dataset in terms of diversity, we conducted a simple analysis on the question type.
As shown in Table \ref{tb:question-type}, VQAG with different configurations corresponds to different distributions of question types, while more than 50\% of the questions in the other datasets contain ``what''.
Among the QAG methods, this point is unique to VQAG.

\begin{table}[tbp]
\setlength{\tabcolsep}{2pt}
\centering
\small
\begin{tabular}{c|rrrrrrr}
\toprule
               Dataset & \multicolumn{1}{c}{what} & \multicolumn{1}{c}{how} & \multicolumn{1}{c}{who} & \multicolumn{1}{c}{which} & \multicolumn{1}{c}{when} & \multicolumn{1}{c}{where} & \multicolumn{1}{c}{why} \\
\midrule
\squaddu{train} & \underline{58.3}                     & 10.4                    & 10.3                    & 6.7                       & 6.7                      & 4.2                       & 1.5                        \\
\squaddu{test} & \underline{56.5}                     & 12.1                    & 11.5                    & 8.6                       & 6.0                      & 3.8                       & 0.8                        \\
HarQG          & \underline{61.3}                     & 7.8                     & 13.8                    & 0.7                       & 10.1                     & 5.8                       & 0.5                      \\
SemQG          & \underline{71.1}                     & 8.1                     & 12.8                    & 1.3                       & 3.6                      & 2.7                       & 0.2                       \\
InfoHCVAE      & \underline{77.1}                     & 6.6                     & 5.0                     & 1.6                       & 5.6                      & 3.3                       & 0.5                      \\
VQAG           &                          &                         &                         &                           &                          &                           &                       \\
~~ ${\mathcal{D}_{5,5}}$           & 36.6                     & \underline{54.9}                    & 4.9                     & 0.5                       & 0.3                      & 0.5                       & 2.3                    \\
~~ ${\mathcal{D}_{5,20}}$          & 9.5                      & 35.5                    & 3.6                     & \underline{49.2}                      & 1.2                      & 0.9                       & 0.0                     \\
~~ ${\mathcal{D}_{20,20}}$         & 28.2                     & \underline{36.7}                    & 6.3                     & 23.2                      & 0.2                      & 1.6                       & 3.9                    \\
&&&&&&& \hfill (\%)\\
\bottomrule
\end{tabular}
\caption{Percentages (\%) of each question type in each dataset. The largest number in each line is \underline{underlined}. VQAG is less likely to contain ``what'' and more likely to contain ``which'' and ``how'' than other data sets.}
\label{tb:question-type}
\end{table}


\section{Discussion and Conclusion}
\label{sec:conclusion}
We presented a variational QAG model, incorporating two independent latent random variables.
We showed that an explicit KL control can enable our model to significantly improve the diversity of QA pairs.
Our synthetic datasets were shown to be noisy in terms of the grammaticality and answerability of questions, but effective in improving the QA performance in the in- and out-of-distribution test sets.
While our synthetic datasets are noisy, they may unintentionally improve the robustness to the noise that can occur in real applications.
However, we should pay attention to the negative effect of using our noisy dataset.
For example, the lack of the answerability of our synthetic questions may lead to the poor performance in handling unanswerable questions such as SQuAD v2.0.

Moreover, the QAG methods led to improvements in most of the 12 challenge sets while being agnostic to the target distributions during training.
We need to pursue such a target-unaware method to improve the robustness of QA models, because it is quite difficult for developers to know the types of questions a QA model cannot handle in advance.

In summary, our experimental results showed that the diversity of QA datasets plays a non-negligible role in improving its robustness, which can be improved with QAG.
We will consider using unlabeled documents in other domains to further improve the robustness to other domain corpora in our future study.

\section*{Acknowledgements}
We would like to thank the anonymous reviewers for their valuable comments.
We would also like to thank the members of Aizawa Lab for their helpful discussions.
This work was supported by NEDO SIP-2 ``Big-data and AI-enabled Cyberspace Technologies'' and JSPS KAKENHI Grant Number 20K23335.

\bibliographystyle{acl_natbib}
\bibliography{acl2021}


\appendix
\section{Derivations of the Variational Lower Bound}
\label{app:derivation}

\noindent The variational lower bound, Eq. \ref{eq:modified-elbo}, without the KL control is derived as follows:

\begin{eqnarray*}
\label{eq:vqag-flat-elbo-deriv}
\lefteqn{\log p_{\theta}(q, a|c)}\\
& = & \mathbb{E}_{z,y \sim q_{\phi}(z,y|q,a,c)} \left[
\log p_{\theta}(q, a|c) \right] \\
& = & \mathbb{E}_{z,y} \left[
\log  \frac{p_{\theta}(q,a|z,y,c)p_{\theta}(z,y|c)}{p_{\theta}(z,y|q,a,c)}
\right]\\
& = & \mathbb{E}_{z,y} \left[
\log \frac{p_{\theta}(q,a|z,y,c)p_{\theta}(z,y|c)}{p_{\theta}(z,y|q,a,c)}\right.\\
& & \left. + \log \frac{q_{\phi}(z,y|q,a,c)}{q_{\phi}(z,y|q,a,c)}
\right]\\
& = & \mathbb{E}_{z,y} \left[
\log \frac{p_{\theta}(q|y,a,c)p_{\theta}(y|c)}{p_{\theta}(y|q,c)}\right.\\
& & +  \log \frac{p_{\theta}(a|z,c)p_{\theta}(z|c)}{p_{\theta}(z|a,c)} \\
& & \left. + \log \frac{q_{\phi}(y|q,c)}{q_{\phi}(y|q,c)} + \log \frac{q_{\phi}(z|a,c)}{q_{\phi}(z|a,c)}
\right]\\
& = & \mathbb{E}_{z,y} \left[
\log p_{\theta}(q|y,a,c) + \log p_{\theta}(a|z,c) \right.\\
& &\left. + \log \frac{p_{\theta}(y|c)}{q_{\phi}(y|q,c)}
+ \log \frac{q_{\phi}(y|q,c)}{p_{\theta}(y|q,c)}\right.\\
& & \left. + \log \frac{p_{\theta}(z|c)}{q_{\phi}(z|a,c)}
+ \log \frac{q_{\phi}(z|a,c)}{p_{\theta}(z|a,c)}
\right]\\
& = & \mathbb{E}_{z,y} \left[ \log p_{\theta}(q|y,a,c) + \log p_{\theta}(a|z,c) \right]\\
& & - D_{\rm KL}(q_{\phi}(y|q,c)||p_{\theta}(y|c))\\
& & + D_{\rm KL}(q_{\phi}(y|q,c)||p_{\theta}(y|q,c))\\
& & - D_{\rm KL}(q_{\phi}(z|a,c)||p_{\theta}(z|c))\\
& & + D_{\rm KL}(q_{\phi}(z|a,c)||p_{\theta}(z|a,c))\\
& \geq & \mathbb{E}_{z,y} \left[ \log p_{\theta}(q|y,a,c) + \log p_{\theta}(a|z,c) \right]\\
& & - D_{\rm KL}(q_{\phi}(y|q,c)||p_{\theta}(y|c))\\
& & - D_{\rm KL}(q_{\phi}(z|a,c)||p_{\theta}(z|c)).\\
\end{eqnarray*}

\section{Distribution Modeling Capacity}
\label{app:modeling-capacity}
We originally developed a QA pair modeling task to evaluate and compare QA pair generation models.
We compared models based on the probability they assigned to the ground truth QA pairs.
We used the negative log likelihood (NLL) of QA pairs as the metric, namely, $-\log p(q, a|c)$.
Since variational models can not directly compute NLL, we estimate NLL with importance sampling.
We also estimate each term in decomposed NLL, i.e.,${\rm NLL_a} = -\log p(a|c)$ and ${\rm NLL_q} = -\log p(q|a,c)$.
The better a model performs in this task, the better it fits the test set.
As a baseline, to assess the effect of incorporating latent random variables, we implemented a pipeline model similar to \citet{Subramanian18} using a deterministic pointer network.

\noindent\textbf{Result}~ Table \ref{tb:nll} shows the result of QA pair modeling.
First, our models with ${\rm C=0}$ are superior to the pipeline model, which means that introducing latent random variables aid QA pair modeling capacity.
However, the KL terms converge to zero with ${\rm C=0}$.
When we set $C>0$, KL values are greater than 0, which implies that latent variables have non-trivial information about questions and answers.
Also, we observe that the target value of KL $C$ can control the KL values, showing the potential to avoid the posterior collapse issue.
\begin{table}[htbp]
\setlength{\tabcolsep}{4pt}
\small
\centering
\begin{tabular}{lrrrrr}
\toprule
 & NLL  & ${\rm NLL}_a$ & ${\rm NLL}_q$ & $D_{{\rm KL}_z}$ & $D_{{\rm KL}_y}$ \\
\midrule
\multicolumn{1}{l}{Pipeline} & 36.26 & 3.99 & 32.50 & - & - \\
\multicolumn{4}{l}{VQAG} \\
~~${\rm C=0}$ & \textbf{34.46} & 4.46 & 30.00 & 0.027 & 0.036 \\
~~${\rm C=5}$ & 37.00 & 5.15 & 31.51 & 4.862 & 4.745 \\
~~${\rm C=20}$ & 59.66 & 14.38 & 43.56 & 17.821 & 17.038 \\
~~${\rm C=100}$ & 199.43 & 81.01 & 112.37 & 92.342 & 91.635 \\
 \bottomrule
\end{tabular}
\caption{QA pair modeling capacity measured on \squaddu{test}. We used the same value $C$ for the target values of KL ${\rm C_a}$ and ${\rm C_q}$ for simplicity. NLL: negative log likelihood of QA pairs.
${\rm NLL}_a$ (${\rm NLL}_q$): NLL of answers (questions).
$D_{{\rm KL}_z}$ and $D_{{\rm KL}_y}$ are Kullback--Leibler divergence in Eq. \ref{eq:modified-elbo}.
NLL for our models are estimated with importance sampling using 300 samples.}
\label{tb:nll}
\end{table}

\section{Information Theoretic Interpretation of the KL control}
\label{app:interpretation}
When training our models, we miximized the variational lower bound in Eq. \ref{eq:modified-elbo} is averaged over the training samples.
In other words, the expectation with respect to the data distribution is maximized.
In the ideal case, the approximated posterior $q_\phi(z|a, c)$ is equal to the true posterior $p_\theta(z|a, c)$.
Then, the expectation of the KL terms with respect to the data distribution is equivalent to the conditional mutual information $I(Z; A|C)$.

Mathematically, when the approximated posterior $q_\phi$ is equal to the true posterior $p_\theta$, the expectation of the KL terms in Eq. \ref{eq:modified-elbo} with respect to the data distribution is:
\begin{eqnarray*}
\lefteqn{\mathbb{E}_{p(q, a, c)} [D_{\rm KL}(p(z|a, c)||p(z|c))]}\\
& = & \sum_{a, c}  p(a, c) \int_z p(z|a, c) \log \frac{p(z|a, c)}{p(z|c)} dz\\
& = & \sum_{a,c} \int_z p(a, c, z) \log \frac{p(a, z|c)}{p(z|c)p(a|c)} dz\\
& = & I(Z; A|C).\\
\end{eqnarray*}
\noindent Thus, controlling the KL term in Eq. \ref{eq:modified-elbo} is equivalent to controlling the conditional mutual information.
The same is true for question $I(Y; Q|C)$.

\section{Model Architechture}
\label{app:architechture}
\paragraph{Prior and Posterior Distribution}
Following \citet{cvae-dialog}, we hypothesized that the prior and posterior distributions of the latent variables follow multivariate Gaussian distributions with diagonal covariance.
The distributions are described as follows:
\begin{align}
z|a, c & \sim \mathcal{N}(\mu_{post_Z}, diag(\sigma^2_{post_Z})) \\
z|c & \sim \mathcal{N}(\mu_{prior_Z}, diag(\sigma^2_{prior_Z})) \\
y|q, c & \sim \mathcal{N}(\mu_{post_Y}, diag(\sigma^2_{post_Y})) \\
y|c & \sim \mathcal{N}(\mu_{prior_Y}, diag(\sigma^2_{prior_Y}). 
\end{align}
\noindent The prior and posterior distributions of the latent variables, $z$ and $y$, are computed as follows:
\begin{align}
& \left[\begin{array}{c}
\mu_{post_Z} \\ \log (\sigma^2_{post_Z})
\end{array}\right] = W_{post_Z} \left[ \begin{array}{c}
h^C \\ h^A
\end{array} \right] + b_{post_Z} \\
& \left[\begin{array}{c}
\mu_{prior_Z} \\ \log (\sigma^2_{prior_Z})
\end{array}\right] = W_{prior_Z} h^C + b_{prior_Z} \\
& \left[\begin{array}{c}
\mu_{post_Y} \\ \log (\sigma^2_{post_Y})
\end{array}\right] = W_{post_Y} \left[ \begin{array}{c}
h^C \\ h^Q
\end{array} \right] + b_{post_Y} \\
& \left[\begin{array}{c}
\mu_{prior_Y} \\ \log (\sigma^2_{prior_Y})
\end{array}\right] = W_{prior_Y} h^C + b_{prior_Y}. 
\end{align}
\noindent Then, latent variable $z$ (and $y$) is obtained using the reparameterization trick \cite{vae}: $z = \mu + \sigma \odot \epsilon$, where $\odot$ represents the Hadamard product, and $\epsilon \sim \mathcal{N}(0, I)$.
Then, $z$ and $y$ is passed to the AE and QG models, respectively.

\paragraph{Answer Extraction Model}
We regard answer extraction as two-step sequential decoding, i.e.,
\begin{align}
\label{eq:ae}
p(a|c)=p(c_{end}|c_{start}, c)p(c_{start}|c),
\end{align}
which predicts the start and end positions of an answer span in this order.
For AE, we modify a pointer network \cite{ptrnet} to take into account the initial hidden state $h_0^{AE} = W_1 z + b_1$, which in the end diversify AE by learning the mappings from $z$ to $a$.
The decoding process is as follows:
\begin{align}
& h^{IN}_{i} = \left\{
\begin{array}{ll}
e(\Rightarrow) & {\rm if}~ i=1 \\
H^C_{t_{i-1}} & {\rm if}~ i=2
\end{array}
\right.
 \\
& h_i^{AE} = {\rm LSTM}(h_{i-1}^{AE}, h^{IN}_i) \\
& u^{AE}_{ij} = (v^{AE})^T {\rm tanh}(W_2 H^C_j + W_3 h_i^{AE} + b_2) \\
& p(c_{t_i}|c_{t_{i-1}}, c) = {\rm softmax}(u_i) 
\label{eq:vpn-2nd}
\end{align}
where $1\leq i \leq 2$, $1\leq j\leq L_C$, $h_i^{AE}$ is the hidden state vector of the LSTM, $h^{IN}_i$ is the $i$-th input, $t_i$ denotes the start ($i$=1) or end ($i$=2) positions in $c$, and $v$, $W_n$ and $b_n$ are learnable parameters.
We learn the embedding of the special token ``$\Rightarrow$'' as the initial input $h^{IN}_1$.

When we used the embedding vector $e_{t_i}$ as $h^{IN}_{i+1}$, instead of $H^C_{t_i}$, following \citet{Subramanian18}, we observed that the extracted spans tended to be long and unreasonable.
We assume that this is because the decoder cannot get the positional information from the input in each step.

\paragraph{Question Generation Model}
For QG, we modify an LSTM decoder with attention and copying mechanisms to take the initial hidden state $h_0^{QG} = W_4 y + b_3$ as input to diversify QG.
In detail, at each time step, the probability distribution of generating words from vocabulary using attention \cite{Bahdanau14} is computed as:
\begin{align}
& h_i^{QG} = {\rm LSTM}(h_{i-1}^{QG}, q_{t-1})  \\
& u^{att}_{ij} = (v^{att})^T {\rm tanh}(W_5 h_i^{QG} + W_6 H^{CA}_j + b_4) \\
& a^{att}_i = {\rm softmax} (u^{att}_i) \\
& \hat{h}_i =  \textstyle\sum_j a^{att}_{ij} H^{CA}_j \\
& \Tilde{h}_i = {\rm tanh} (W_7([\hat{h}_i; h_i^{QG}] + b_5)) \\
& P_{vocab} = {\rm softmax} (W_8(\Tilde{h}_i) + b_6), 
\end{align}
\noindent and the probability distributions of copying \cite{Gulcehre16,Gu16} from context are computed as:
\begin{align}
& u^{copy}_{ij} = (v^{copy})^T {\rm tanh}(W_9 h_i^{QG} + W_{10} H^{CA}_j + b_7) \\
& a^{copy}_i = {\rm softmax}(u^{copy}_i) 
\end{align}
\noindent Accordingly, the probability of outputting $q_i$ is:
\begin{align}
& p_{g} = \sigma (W_{11} h_i^{QG}) \\
& p(q_i|q_{1:i-1}, a, c)\\
& = p_{g} P_{vocab} (q_i)  + (1 - p_{g}) \textstyle\sum_{j:c_j={q_i}} a^{copy}_{ij} 
\end{align}
\noindent where $\sigma$ is the sigmoid function.

\section{Training Details}
\label{app:training-details}
We use pretrained GloVe \cite{glove} vectors with 300 dimensions and freeze them during training.
The pretrained word embeddings were shared by the input layer of the context encoder, the input and output layers of the question decoder.
The vocabulary has most frequent 45k words in our training set.
The dimension of character-level embedding vectors is 32.
The number of windows is 100.
The dimension of hidden vectors is 300.
The dimension of latent variables is 200.
All LSTMs used in this paper have one layer. We used Adam \cite{kingma2014adam} for optimization with initial learning rate 0.001. All the parameters were initialized with Xavier Initialization \cite{xavier}.
Models were trained for 16 epochs with a batch size of 32.
We used a dropout \cite{dropout} rate of 0.2 for all the LSTM layers and attention modules.

\section{Answer Extraction and Question Generation}
\label{app:ae-qg}
Tables \ref{tb:ae-detail} and \ref{tb:qg-detail} show the detailed results of AE and QG.
Various values of ${\rm C_a}$ and ${\rm C_q}$ are explored.

\begin{table}[htbp]
\setlength{\tabcolsep}{2pt}
\small
\centering
\begin{tabular}{lrrlrrlr}
\toprule
\multicolumn{1}{c}{} & \multicolumn{5}{c}{Relevance} & & \multicolumn{1}{l}{Diversity} \\
\cmidrule{2-6}\cmidrule{8-8}
\multicolumn{1}{l}{} & \multicolumn{2}{c}{Precision} &  & \multicolumn{2}{c}{Recall} & & \multicolumn{1}{c}{\multirow{2}{*}{Dist}} \\
\cmidrule{2-3}\cmidrule{5-6}
\multicolumn{1}{c}{} & \multicolumn{1}{c}{Prop.}  & \multicolumn{1}{c}{Exact} &  & \multicolumn{1}{c}{Prop.} & \multicolumn{1}{c}{Exact} &  &  \\
\midrule
NER & 34.44 & 19.61 &  & 64.60 & 45.39 &  & 30.0k \\
BiLSTM-CRF & 45.96 &  33.90 &  & 41.05 &  28.37 &  & - \\
InfoHCVAE & 31.59 & 16.18 &  & 78.75 & 59.32 &  & 70.1k\\
\midrule
VQAG & \multicolumn{1}{l}{} & \multicolumn{1}{l}{} &  & \multicolumn{1}{l}{} & \multicolumn{1}{l}{} &  & \multicolumn{1}{l}{} \\
~~${\rm C_a=0}$ & \textbf{58.39} &  \textbf{47.15} &  & 21.82 & 16.38 &  & 3.1k \\
~~${\rm C_a=3}$ & 34.09 & 19.22 &  & 78.94 & 59.09 &  & 47.5k \\
~~${\rm C_a=5}$ & 30.16 & 13.41 &  & \textbf{83.13} & \textbf{60.88} &  & 71.2k \\
~~${\rm C_a=10}$ & 26.17 & 8.83 &  & 79.70 & 53.02 &  & 92.3k \\
~~${\rm C_a=15}$ & 22.42 & 6.11 &  & 76.18 & 44.80 &  & 99.9k \\
~~${\rm C_a=20}$ & 21.95 & 5.75 &  & 72.26 & 42.15 &  & \textbf{103.3k} \\
~~${\rm C_a=25}$ & 21.60 & 5.37 &  & 71.55 & 40.48 &  & 101.6k \\
~~${\rm C_a=30}$ & 23.88 & 6.75 &  & 74.08 & 44.59 &  & 99.5k \\
~~${\rm C_a=40}$ & 24.58 & 7.90 &  & 74.86 & 43.33 &  & 88.1k \\
~~${\rm C_a=50}$ & 25.05 & 7.83 &  & 76.56 & 44.67 &  & 88.9k \\
~~${\rm C_a=100}$ & 23.32 & 7.48 &  & 71.74 & 39.70 &  & 84.6k \\
\bottomrule
\end{tabular}
\caption{Detailed results of AE on \squaddu{test}.}
\label{tb:ae-detail}
\end{table}

\begin{table*}[htbp]
\setlength{\tabcolsep}{4pt}
\small
\centering
\begin{tabular}{lrrrrrrrrrrr}
\toprule
 & \multicolumn{6}{c}{Relevance} &  & \multicolumn{4}{c}{Diversity} \\
 \cmidrule{2-7}\cmidrule{9-12}
 & \multicolumn{1}{c}{B1} & \multicolumn{1}{c}{B2} & \multicolumn{1}{c}{B3} & \multicolumn{1}{c}{B4} & \multicolumn{1}{c}{ME} & \multicolumn{1}{c}{RL} & \multicolumn{1}{c}{Token} & \multicolumn{1}{c}{D1} & \multicolumn{1}{c}{D2} & \multicolumn{1}{c}{E4} & \multicolumn{1}{c}{SB4} \\
\midrule
\citet{Zhang19} & 48.59 & 32.83 & 24.21 & 18.40 & 24.86 & 46.66 & 133.8k & 10.2k & 46.4k & 15.78 & - \\
\bottomrule
\\
\toprule
 & \multicolumn{1}{c}{B1-R} & \multicolumn{1}{c}{B2-R} & \multicolumn{1}{c}{B3-R} & \multicolumn{1}{c}{B4-R} & \multicolumn{1}{c}{ME-R} & \multicolumn{1}{c}{RL-R} & \multicolumn{1}{c}{Token} & \multicolumn{1}{c}{D1} & \multicolumn{1}{c}{D2} & \multicolumn{1}{c}{E4} & \multicolumn{1}{c}{SB4} \\
\midrule
\citet{Zhang19}
& \textbf{62.32} & \textbf{47.77} & \textbf{37.96} & \textbf{30.05} & \textbf{36.77} & \textbf{62.87} & 7.0M & 15.8k & 218.9k & 18.28 & 91.44 \\
\midrule
\multicolumn{1}{l}{VQAG} &  &  &  &  &  &  &  &  &  &  &  \\
~~${\rm C_q=0}$ & 35.57 & 18.75 & 10.79 & 6.35 & 18.31 & 33.92  & 7.6M & 14.4k & 155.3k & 17.33 & 97.61 \\
~~${\rm C_q=3}$ & 44.05 & 26.74 & 16.08 & 9.26 & 24.61 & 44.10  & 9.0M & 17.8k & 394.2k & 19.14 & 85.88 \\
~~${\rm C_q=5}$ & 44.19 & 27.09 & 16.33 & 9.71 & 25.84 & 45.18  & 11.5M & 19.0k & 481.1k & 19.71 & 82.59 \\
~~${\rm C_q=10}$ & 44.00 & 27.15 & 16.78 & 10.24 & 25.64 & 44.78  & 10.2M &  18.8k & 461.5k & 19.69 & 80.39 \\
~~${\rm C_q=15}$ & 45.23 & 27.91 & 16.67 & 10.11 & 26.12 & 45.41  & 11.3M & 19.5k & 381.5k & 19.40 & 84.56 \\
~~${\rm C_q=20}$ & 48.19 & 32.87 & 22.96 & 14.94 & 25.29 & 48.26  & 4.9M & 22.4k & 549.2k & 19.72 & 44.41 \\
~~${\rm C_q=25}$ & 47.20 & 31.16 & 21.15 & 13.66 & 25.30 & 45.97  & 6.8M & 22.3k & 706.9k & \textbf{20.34} & 47.00 \\
~~${\rm C_q=30}$ & 47.96 & 31.69 & 21.26 & 13.83 & 24.95 & 47.07  & 7.3M & \textbf{22.9k} & \textbf{732.8k} & 18.54 & 50.32 \\
~~${\rm C_q=40}$ & 46.31 & 31.29 & 21.52 & 13.94 & 23.73 & 46.46  & 5.4M & 21.0k & 487.8k & 19.39 & 55.95 \\
~~${\rm C_q=50}$ & 43.92 & 25.95 & 15.54 & 9.61 & 23.61 & 43.18  & 10.8M & 22.2k & 527.2k & 19.29 & 73.78 \\
~~${\rm C_q=100}$ & 35.22 & 19.88 & 13.25 & 9.20 & 22.27 & 37.55 & 8.2M & 22.1k & 508.8k & 19.74 & \textbf{44.22} \\
\bottomrule
\end{tabular}
\caption{Detailed results of answer-aware QG on \squaddu{test}. Paragraph-level contexts and answer spans are used as input. The baseline model is ELMo+QPP\&QAP \cite{Zhang19} with diverse beam search \cite{divbeam} with a beam size 50. Bn: BLEU-n, ME: METEOR, RL: ROUGE-L, Token: the total number of the generated words, Dn: Dist-n, E4: Ent-4 (entropy of 4-grams), SB4: Self-BLEU-4. ``-R" represents recall. (e.g. B1-R is the recall of B1.) One question per answer-context pair is evaluated in the upper part, while 50 questions per answer-context pair are evaluated in the lower part to assess their diversity.}
\label{tb:qg-detail}
\end{table*}

\section{Latent Interpolation}
\label{app:latent-interpolation}
Table \ref{tb:latent-interpolation} shows the latent interpolation between two ground-truth QA pairs using VQAG with $({\rm C_a}, {\rm C_q})=(5, 20)$.
This result shows that $z$ controls answer and $y$ controls question.

\begin{table*}[h!]
\centering
\small
\begin{tabular}{p{0.3cm}|p{2.6cm}|p{2.6cm}|p{2.6cm}|p{2.6cm}|p{2.6cm}} 
 & \cellcolor[rgb]{0.0, 0.9, 0.9} $z_1$  & \cellcolor[rgb]{0.0, 0.8, 0.9} $z_2$  & \cellcolor[rgb]{0.0, 0.7, 0.9} $z_3$  & \cellcolor[rgb]{0.0, 0.6, 0.9} $z_4$  & \cellcolor[rgb]{0.0, 0.5, 0.9} $z_5$\\ \hline 
 \cellcolor[rgb]{1.0, 0.9, 0.0} $y_1$ & \textbf{in what city and state did beyonce   grow up ?}---\textbf{houston , texas} & how do competitions performed a child child ?---dancing & the american singer born what american singer ?---songwriter & how did beyoncé dobruja to ?---dangerously in love & how did beyoncé album album ?---dangerously in love\\ \hline 
 \cellcolor[rgb]{1.0, 0.8, 0.0} $y_2$ & the album born and raised ?---houston , texas & how do competitions enovid ?---dancing & how is actress - carter ?---songwriter & how did beyoncé 's album album ?---dangerously in love & how did beyoncé album album ?---dangerously in love\\ \hline 
 \cellcolor[rgb]{1.0, 0.7, 0.0} $y_3$ & the album born and raised ?---houston , texas & how do competitions performed a child child ?---dancing & the american singer born what american singer ?---songwriter & how did beyoncé dobruja to ?---dangerously in love & how did beyoncé dobruja to ?---dangerously in love\\ \hline
 \cellcolor[rgb]{1.0, 0.6, 0.0} $y_4$ & the album born and raised ?---houston , texas & how many competitions does texas child perform ?---dancing & the american singer born what american singer ?---songwriter & how did beyoncé dobruja to ?---dangerously in love & how did beyoncé dobruja to ?---dangerously in love\\ \hline 
 \cellcolor[rgb]{1.0, 0.5, 0.0} $y_5$ & the album born and raised ?---houston , texas & how many competitions did texas child perform ?---dancing & the american singer born what american singer ?---songwriter & how did beyoncé dobruja to ?---dangerously in love & \textbf{what was the name of beyoncé 's first solo album ?}---\textbf{dangerously in love}\\ \hline 
\end{tabular}
\caption{Latent interpolation with VQAG with $({\rm C_a}, {\rm C_q})=(5, 20)$. The samples in the upper left and lower right are the ground truth QA pairs from the same paragraph as Table \ref{tb:heatmap}. The linearly interpolated samples show how our generative model learns mapping from latent space to QA pairs.}
\label{tb:latent-interpolation}
\end{table*}

\section{Human Evaluation}
\label{app:human}
We conducted human evaluation to assess the quality of QA pairs by asking the following questions.

\begin{enumerate}
\setlength{\parskip}{0cm}
\setlength{\itemsep}{0cm}
\item \textbf{Is the question well-formed in itself?} The workers are asked to select \textit{yes} if a given question is both grammatical and meaningful. The workers select \textit{understandable} if a question is not grammatical but meaningful.
\item \textbf{Is the question relevant to the passage?} This is to check whether a question is relevant to the content of a passage.
\item \textbf{Is the answer a correct answer to the question?} If a given answer partially overlaps with the true answer in a passage, the workers select \textit{partially}.
\item \textbf{Is the meaning of the answer in itself related to the main topic of the passage?} This is to check the importance of an answer. We designed this question to assess the question-worthiness of an answer.
\end{enumerate}

Each triple is evaluated by three crowdworkers. Each task costs 0.08 USD.

\end{document}